# WebMRIQC: A Web-Based Implementation of MRIQC for Accessible MRI Image Quality Assessment in Resource-Constrained Settings

*Philip Nkwam[1], Ifeoluwa Oladeji[6], Sekinat Zurakat-Aderibigbe[1], Jasmine Cakmak[2], Harrison Aduluwa[2], Confidence Raymond[2], Cliff Mokua[3], Abdulrazaq Zubair[4], Daniel Champanda[5], Tolulope Olusuyi[6], Maruf Adewole[6], Udunna Anazodo[2,6]*

[1]Department of Radiography, Faculty of Health Professions, College of Medicine, University of Lagos, Nigeria.

[2]Department of Neurology and Neurosurgery, Montreal Neurological Institute, McGill University, Montreal, Canada.

[3]Jomo Kenyatta University of Agriculture and Technology, Kenya

[4]Federal University of Health Sciences, Azare, Nigeria

[5]Radiology Department , Muhimbili Orthopaedic Institute, Dar es salaam, Tanzania

[6]Medical Artificial Intelligence Laboratory, Crestview Radiology Ltd., Lagos, Nigeria.

*Philipnkwam@gmail.com*

**Abstract.** Reliable quality control (QC) of magnetic resonance imaging (MRI) is essential for reliable diagnostic neuroimaging, yet standard manual assessment is subjective and time-consuming. MRIQC has established standardized automated extraction of image-quality metrics (IQMs), but its reliance on local computational imaging skills and capacity including high-performance computing, limits its adoption in resource-constrained settings (RCS). We present WebMRIQC (webmriqc.mailab.io), an open-source browser-based platform that wraps the validated MRIQC engine behind a zero-installation web interface. WebMRIQC automates the DICOM-to-BIDS conversion of de-identified MRI scans, executes the unmodified containerized MRIQC pipeline on a shared compute node governed by a fair-share job queue, and returns an interactive in-browser dashboard. The dashboard grounds every IQM in published quality thresholds, benchmarks each scan against the normative distribution of high-resource open datasets, and supports cross-site multicentre implementation of optimized scan protocols in RCS.We describe the system architecture and a validation framework establishing measurement equivalence between WebMRIQC and native MRIQC across thirteen IQMs on the BraTS-Africa and BraTS 2021 datasets. Preliminary results indicate strong agreement for contrast-, signal and noise-based metrics, demonstrating that web-based implementation lowers the barrier to standardized MRI QC and provides a foundation for harmonized, regionally adapted quality benchmarks across RCS imaging sites. The code is publicly available here *https://github.com/CAMERA-MRI/WebMRIqc*.

**Keywords:** MRI quality control · MRIQC · Image quality metrics · Low-resource settings · Harmonization · Reproducibility. neuroimaging.

## 1 Introduction

Brain Magnetic resonance imaging (MRI) underpins a growing share of clinical decision-making and neuroscience research. The reliability of any downstream analysis, however, is bounded by the quality of the acquired images, influenced by motion, low signal-to-noise ratio, intensity non-uniformity, and reconstruction artefacts. All of which can obscure genuine biological signals. Quality control (QC) is therefore a prerequisite for reliable diagnostic imaging, yet in routine clinical practice it remains largely subjectively evaluated and challenging to standardize across operators, scanners, sites, and practice settings [1].

MRIQC [2] addressed this gap by providing an open-source pipeline that computes a comprehensive set of image-quality metrics (IQMs) and renders visual reports for anatomical and functional MRI [2]. By transforming subjective visual inspection into reproducible quantitative descriptors derived from brain images, MRIQC has become a de facto quantitative QC approach in neuroimaging research settings with specific applications in cross-site quality assessments for MRI harmonization and test-retest reliability [3-4]. Despite this, the operational requirements of MRIQC present a substantial barrier in resource-constrained settings. It requires local installation of software libraries , organization of data into the Brain Imaging Data Structure (BIDS) [5], and access to multi-core, high-memory computing infrastructure. Each of these present a non-trivial obstacle in resource-constrained settings, where infrastructure and skilled personnel for open source image computing are scarce. This disproportionately affects low-and-middle-income countries (LMICs), where the need for rigorous, harmonized QC is the greatest. Across LMICs, the dominance of many types of lower field clinical MRI scanners (≤1.5T) including region-specific and exclusive vendor options [6], can limit equitable participation in collaborative interregional and global neuroimaging research as this depends on demonstrable data quality [7]. We argue that the obstacle is one of accessibility rather than methodology, and that the established MRIQC pipeline should be made usable without local installation, image computing expertise, or dedicated hardware.

Here, we present Web-MRIQC, an open-source browser-based platform that delivers the full MRIQC analysis through a zero-installation web interface designed for resource-constrained settings. Our contributions are fourfold. First, Web-MRIQC wraps the unmodified, containerized MRIQC engine to generate IQM reports that are identical to the native MRIQC with no software setup on the user's device. Second, it automates the DICOM-to-BIDS conversion step, a critical image processing initialization process for harmonizing scans from multiple vendors, removing the most common barrier to running BIDS compliant QC, especially on new/region-specific scanners (e.g., United Imaging, Anke, Neusoft, etc). Third, it shares a single high-capacity compute node fairly among many concurrent users through a memory-aware job queue, an arrangement well suited to settings where individual institutions cannot provision their own high-performance computing but can run their analysis simultaneously with others. Fourth, it adds a percentile ranking-based benchmarking layer for validating the quality of locally-adapted acquisition protocols against a normative distribution (established open datasets/established protocols) or for direct cross-site comparisons, to support MRI harmonization. The platform was validated against the native MRIQC implementation using the BraTS 2021 [8] and BraTS-Africa [7] Challenge datasets, to evaluate its QC measurement equivalency and the new benchmarking extension.

## 2 Related Work

***Automated MRI quality assessment.*** Early work established that structural MRI quality could be characterized by quantitative descriptors of the image background and tissue distributions, including the Quality Index proposed by Mortamet et al. [1].

MRIQC consolidated and extended these into a unified, modality-aware framework and demonstrated that IQMs computed at one site could predict expert quality ratings at unseen sites [2,3]. Subsequent analyses have characterized the reliability of these metrics across repeated acquisitions, motivating their use as harmonization targets [4]. WebMRIQC does not modify or reimplement these metrics. Instead, it preserves the validated engine and focuses on access and interpretation.

***Standardization and BIDS Apps.*** The Brain Imaging Data Structure [5] and the BIDS Apps model [9] standardized neuroimaging metadata by introducing DICOM conversion formats for improved interoperability and reproducibility. In practice, however, organizing heterogeneous clinical DICOM exports into valid BIDS format from various scanner manufacturers remains a recurring obstacle, particularly for users without coding experience. Established tools such as dcm2niix [10] convert DICOM to NIfTI, and configurable wrappers organize the result into BIDS. WebMRIQC automates this end-to-end process so users upload raw DICOM MRI data directly for conversion to BIDS.

***Web-based neuroimaging***. Browser-based approaches have been used to scale QC through expert and citizen review, for example by combining crowdsourcing with deep learning to amplify expert quality judgements [11,12]. These efforts demonstrate the value of moving QC into an install-free platform but current approaches target visual rating rather than turnkey quantitative metric computation. WebMRIQC complements this by delivering both visual and fully quantitative install-free MRIQC analysis, including automated preprocessing in a user-friendly web platform.

***Accessible imaging in LMICs.*** Initiatives to advance sustainable MRI access and education in Africa have highlighted infrastructure, training, and harmonization as central challenges [11], and the release of regionally representative datasets such as BraTS-Africa [14] are addressing the under-representation of African populations in benchmark data. WebMRIQC is designed to support these goals operationally, providing a shared QC ecosystem and the normative comparisons needed to define regionally adapted quality benchmarks across LMICs.

## 3 System Design and Methods

### 3.1 *Design Goals*

WebMRIQC was guided by five goals derived from the constraints of resource-limited deployment: (i) open-source zero client-side installation, accessible from common browsers; (ii) fidelity to the native MRIQC and computing IQMs with the unmodified upstream engine rather than a re-implementation; (iii) automation of preprocessing without BIDS expertise for upload of raw de-identified DICOM conversion; (iv) fair sharing of a single compute resource among many users in constrained settings; and (v) data minimization with transient data processing of uploaded data for frugal, efficient, and privacy-aware data management computing. A normative benchmarking approach was added to make reproducible QC metrics actionable for LMIC clinicians, MR technologists, and neuroscience researchers. This will enable inclusion of LMIC brain MRI datasets, especially from African populations, in multi-center studies, such as expansion of open neuroimaging datasets [14] or local implementation of established brain protocols [15]. To support non-expert users, WebMRIQC includes an in-application support feature (shown in Figure 1) that assists with issues encountered during use, most commonly failures of DICOM-to-BIDS conversion caused by non-standard DICOM metadata. This is currently provided as an IT support channel through email contact and an AI-enabled support chatbot (implemented using the Google Gemini API). The chatbot feature is functional and offers interactive, plain-language guidance, but has not yet been fully tested.

### 3.2 *Architecture Overview*

The platform follows a three-tier architecture (Figure 1):

1. The client tier is a single-page web application (front-end) that handles DICOM uploads and all result visualization.

2. The processing tier is a back-end server that validates input, performs automated DICOM-to-BIDS conversion, and schedules work through a fair-share and capacity-bounded job queue.

3. The compute tier executes the containerized MRIQC pipeline and QC benchmarking on a shared high-performance node and returns per-subject IQMs, with visual-report figures to the browser.

All three tiers are packaged within a single reproducible Docker container image [16] and exposed over an encrypted tunnel, such that the uploaded de-identified data are processed transiently and permanently deleted once the results archive has been retrieved.

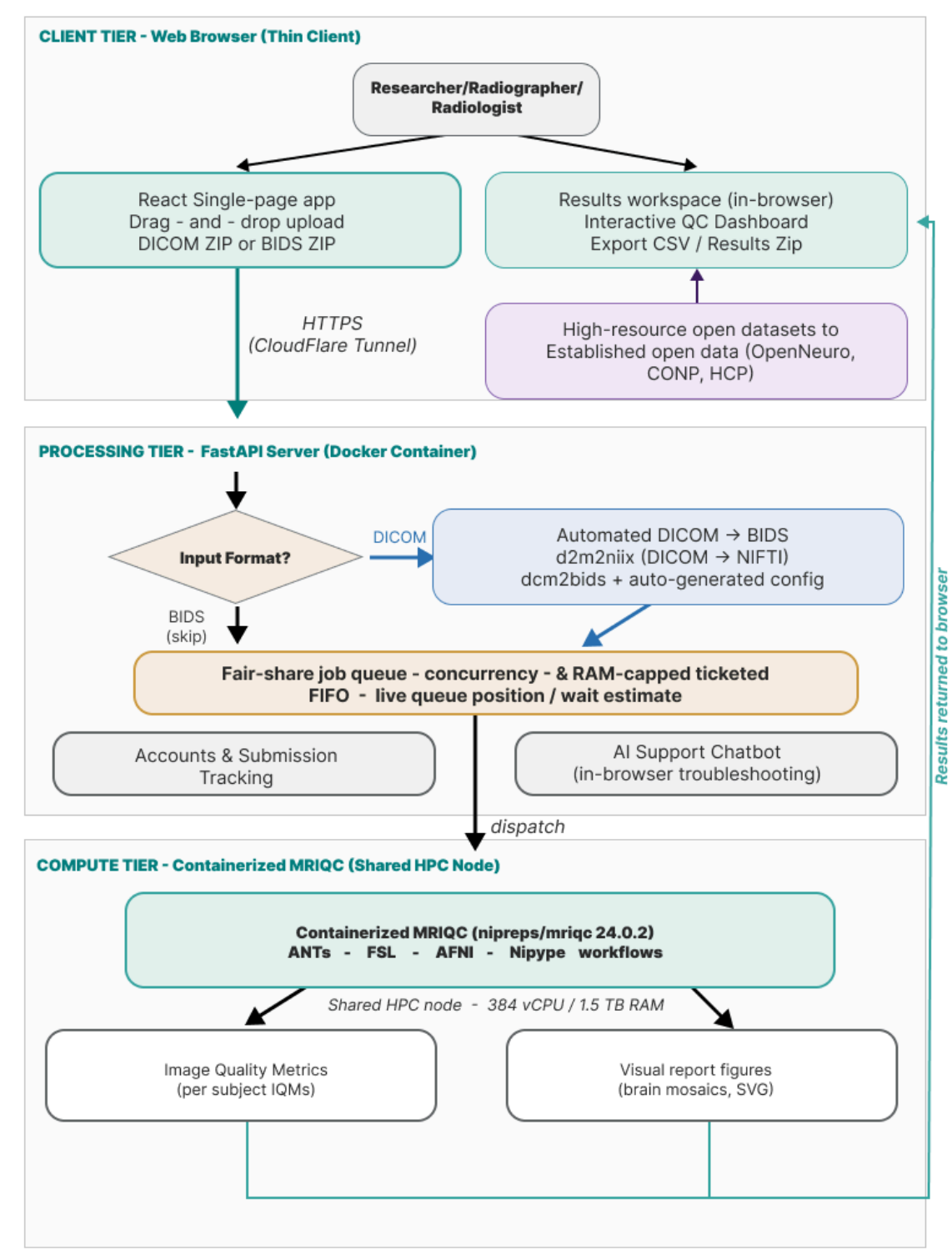

**Figure 1.** WebMRIQC three-tier data flow. A user uploads a DICOM or BIDS archive from the browser (client tier), which is then sent to the server (processing tier) to automatically convert the DICOM to BIDS and schedule the job through a memory-aware fair-share queue. The unmodified containerized MRIQC engine, then runs on a shared high-performance node (compute tier) and returns per-subject IQMs and visual reports to the browser. The results are finally rendered as an interactive dashboard, benchmarked against standard open datasets via percentile ranking, and compared across sites. The uploaded images are deleted after IQMs retrieval, minimizing storage and data governance burden.

### 3.3 *Automated DICOM-to-BIDS Conversion*

Requiring BIDS-formatted input, in our experience, is the single most common reason that users fail to run MRIQC. WebMRIQC therefore accepts either BIDS data or raw DICOM data. For DICOM input, the server converts images to NIfTI using dcm2niix [9] and organizes them into a valid BIDS format using dcm2bids [17] driven by an automatically generated configuration. The conversion stage infers subject and session labels, derives sidecar metadata, resolves common packaging peculiarities (e.g., single nested top-level folder), and writes the dataset description and accompanying files required for BIDS validity. The user then simply uploads a familiar DICOM folder and receives QC results, while users with BIDS datasets bypass the conversion step. This automation operationalizes the BIDS Apps model [8] for users who would otherwise be excluded by the formatting requirement.

### 3.4 *Containerized MRIQC Execution*

The compute tier invokes the official MRIQC container (nipreps/mriqc, version 24.0.2), which bundles MRIQC together with its scientific dependencies, including ANTs, FSL, AFNI, and Nipype-based workflows. Because the engine is the upstream image rather than a re-implementation, the IQMs and visual reports are produced by exactly the same code paths as a local MRIQC run,maintaining measurement equivalence (c.f. Section 4). For each participant, MRIQC writes the IQMs as structured JSON and the mosaic visualizations as scalable vector graphics, both of which are returned to the browser for rendering. Resource allocation per job (CPU cores and memory) is configurable and is bounded by the queue, as described below.

### 3.5 *Fair-Share Job Queue*

A defining constraint of resource-limited settings is that individual institutions rarely own high-performance computers. A viable workaround is to share a well-provisioned node among many sites. WebMRIQC implements a memory-aware, ticketed first-in-first-out queue that caps the number of concurrent MRIQC jobs and the memory committed to them, preventing large jobs from exhausting the node or degrading service for others. The maximum threshold capacity is currently 20 concurrent submissions. Submissions beyond this capacity receive a queue ticket with a live position and an estimated wait, and are promoted automatically as resources become free. This scheduling makes a single shared deployment (a node with 384 virtual CPUs and 1.5 TB of RAM) serve many concurrent users predictably, amortizing infrastructure that no single partner site could justify alone. The node is part of the Neuro Open Science and Global Health Platform (NeurON) physical cluster, located at the Medical Artificial Intelligence Laboratory (MAI Lab), Lagos, Nigeria, a relatively well equipped computing imaging lab in Africa with dedicated computing infrastructure and highly qualified imaging informatics experts [18].

### 3.6 *Interactive and Explainable In-Browser Quality Dashboard*

Results are presented as an interactive dashboard rather than raw files. Each IQM is shown as a card that reports the value, a colour-coded quality rating, and a plain-language description. The rating thresholds are drawn from the published QC literature - for example, the entropy focus criterion for ghosting and blurring [19] and intensity non-uniformity estimated with the N4 bias-field model [20]. Each card links to the primary reference so that non-specialist users can understand what the metric is and why a value is acceptable or not. This extends the native MRIQC IQM descriptions to a readily explainable tool. The standard MRIQC brain image mosaic figures are presented as a browsable gallery, and the complete set of metrics can be exported as a comma-separated file or downloaded as the full results archive.

### 3.7 *Normative Reference and Multi-centre Comparisons*

A metric value is difficult to act on without a frame of reference, particularly for local adaptation of established acquisition protocols or harmonization [21]. WebMRIQC contextualizes each scan against a normative distribution of established open datasets. For each IQM, the platform computes the percentile rank of the user's value within a reference population of openly available anatomical and functional MRI scans curated from neuroimaging repositories (e.g., OpenNeuro, The Canadian Open Neuroscience Platform (CONP)). A scan acquired on a clinical LMIC 1.5 T system, for instance, can thus be quality-assured against standard research-grade acquisitions [21], enabling regionally-adapted benchmarks as local representative reference distributions are assembled. Thus the same percentile ranking approach can be used to assess how well an African site's scans compare to high-resource standards and to regional constraints.

These cross-site comparisons are available on the platform’s multicentre comparison view, in which IQMs from multiple authenticated sites/datasets are tabulated side by side with summary statistics and per-metric quality colour-coding. The view ingests results produced by the platform as well as externally supplied metric tables in either comma- or tab-separated form, so that every site is directly compared against a common reference/baseline. This makes cross-scanner/site differences visible at a glance and provides the substrate for re-defining shared acquisition standards.

### 3.8 *Deployment, Security, and Data Governance*

The application is distributed as a single Docker container image, served over HTTPS through an encrypted tunnel. Consistent with data-minimization and privacy principles [22,23], de-identified uploaded datasets are processed in a transient storage and deleted automatically and permanently after the extracted IQMs and mosaic figure captures are retrieved. Therefore, no imaging data is retained and data governance burden is minimized. The general use of the platform requires account registration for access with a password-protected account that can track the status of their submissions, and view multi-centre comparisons. Password reset and recovery flows store no plain-text secrets, maintaining the privacy expectations appropriate for clinical imaging, while keeping the barrier to entry for research applications at a minimum. WebMRIQC is intended strictly for de-identified data obtained under appropriate institutional approvals. Prior to upload, users are responsible for ensuring that the necessary institutional/ethical review board (IRB) approvals and data-use agreements are in place and that all data have been fully de-identified; these requirements are reinforced at the point of use through the platform's usage guide ("How it Works") and a dedicated FAQ. Consistent with the platform's data-minimization design, uploaded data are transmitted over an encrypted channel, processed in transient storage, and permanently deleted after the results archive is retrieved, so no clinical imaging data are retained.

# 4 Validation Framework and Preliminary Results

The platform was first evaluated by 6 users in Africa to test its functionality and resolve any bugs/issues. Because WebMRIQC executes the unmodified MRIQC engine, our equivalence validation tests whether the IQMs returned through the browser are statistically indistinguishable from those produced by a native MRIQC installation on the same data, and whether processing remains efficient under shared, concurrent use. For each scan, WebMRIQC and native MRIQC IQMs were computed and thirteen core metrics were compared: the coefficient of joint variation (CJV), contrast-to-noise ratio (CNR), entropy focus criterion (EFC), Full Width Half Minimum (FWHM), White-matter-to-maximum-intensity ratio (WM2MAX), Foreground-Background Energy Ratio (FBER) and signal-to-noise ratio (SNR) in cerebrospinal fluid (CSF), grey matter (GM), white matter (WM), and across the whole brain (SNR-total). Native-versus-web IQM differences were assessed using paired t-tests, absolute agreement using the intraclass correlation coefficient ICC(2,1), and Bland-Altman analysis for bias; the relationship between the two implementations was evaluated using Pearson correlation. Significance for these equivalence analyses was set at $p < 0.05$. Separately, for the between-dataset benchmarking comparison of the four core IQMs (BraTS 2021 versus BraTS-Africa; Figure 3), group differences were evaluated using Welch's unpaired t-test, with statistical significance for this test set at $p < 0.0001$. All statistical analyses were conducted using R software (version 4.6.1).

## 4.1 *Datasets*

To validate the platform and evaluate its benchmarking feature, we used 20 T1-weighted MRI scans from the BraTS-Africa dataset [14] and 20 scans from the BraTS 2021 dataset [8]. The BraTS-Africa dataset is Africa's first publicly available benchmarked and multi-center brain MRI data, which provides regionally representative multi-parametric anatomical scans from Sub-Saharan Africa. The data was previously curated following the standard brain tumor segmentation (BraTS) protocol and the image quality was benchmarked to the BraTS 2021 challenge dataset, using a native MRIQC implementation [14]. The data also permits probing of quality metrics stability under a more realistic condition (glioma pathology) for clinical quality assurance.

## 4.2 *Preliminary Results*

The WebMRIQC is publicly accessible from [webmriqc.mailab.io](webmriqc.mailab.io), with the front-end view shown in Figure 2. Initial testing of the platform's functionalities were successfully performed by four radiographers in Nigeria, Tanzania, and Kenya, each uploading one volunteer test scan from their clinical 1.5T MRI systems (Siemens, Philips, and United Imaging), as part of a MRI acquisition skills training course [6]. Two researchers in Nigeria independently evaluated the platform's utility using the BraTS-Africa data. Preliminary evaluation, revealed, no statistically significant differences in any of the thirteen IQMs between the native and web implementations ($p > 0.05$; Table 2). Contrast- and noise-based metrics showed the strongest agreement: CJV, CNR, and EFC each exhibited strong linear correlation ($r \geq 0.8$), with excellent absolute agreement for CJV and CNR and good agreement for EFC and SNR-GM. Agreement was moderate for SNR-CSF, SNR-total, and SNR-WM, mirroring the lower intrinsic reliability previously reported for these particular metrics [4] rather than indicating any platform discrepancy. Table 1 summarizes the preliminary agreement.

## 4.3 *Interactive and Explainable In-Browser Quality Dashboard*

Results are presented as an interactive dashboard rather than raw files. Each IQM is shown as a card that reports the value, a colour-coded quality rating, and a plain-language description. The rating thresholds are drawn from the published QC literature - for example, the entropy focus criterion for ghosting and blurring [19] and intensity non-uniformity estimated with the N4 bias-field model [20]. Each card links to the primary reference so that non-specialist users can understand what the metric is and why a value is acceptable or not. This extends the native MRIQC IQM descriptions to a readily explainable tool. The standard MRIQC brain image mosaic figures are presented as a browsable gallery, and the complete set of metrics can be exported as a comma-separated file or downloaded as the full results archive.

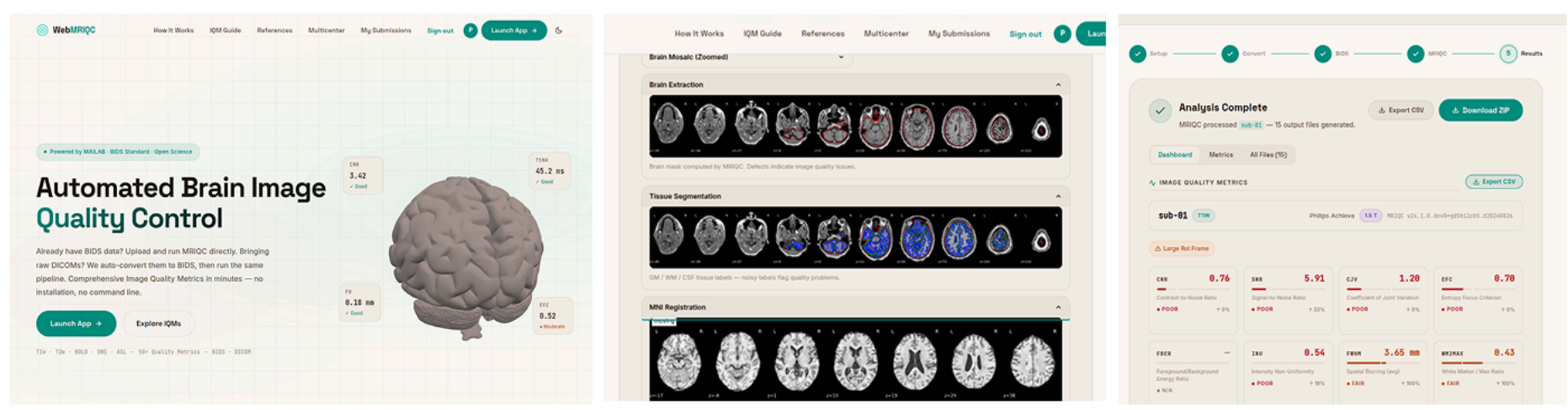

**Figure 2.** A screen capture of WebMRIQC's front end showcasing the homepage and an interactive display of a part of the results.

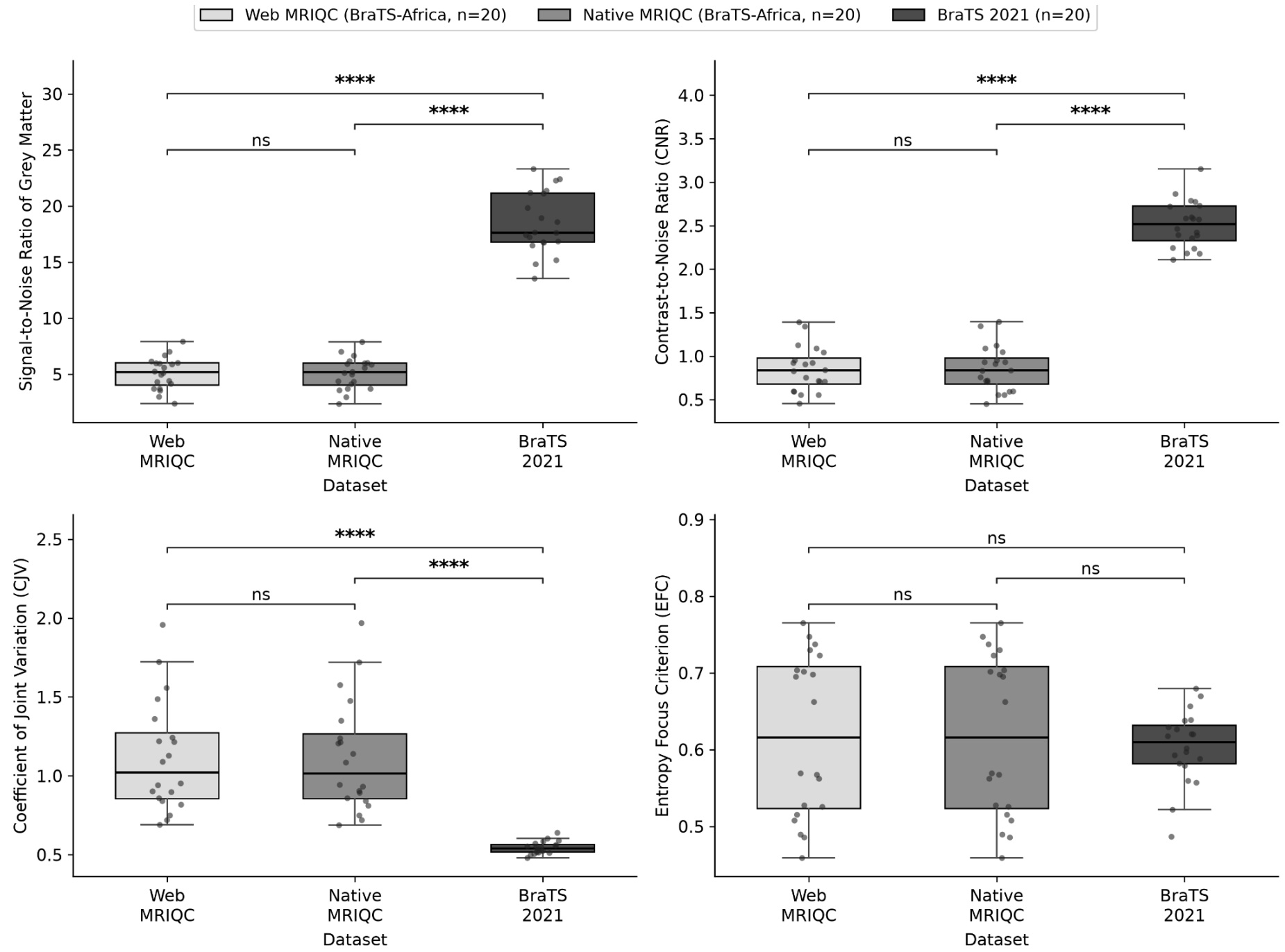

**Figure 3.** Comparison of four core image-quality metrics between the BraTS 2021 benchmark (n = 20) and the BraTS-Africa dataset as measured by WebMRIQC and Native-MRIQC (n = 20). Boxes show the median and interquartile range; whiskers extend to the most extreme values within 1.5×IQR; points show individual subject values. Significance assessed with Welch's unpaired t-test (****p < 0.0001; ns, not significant).

| IQM | Mean ± SD (Native) | Mean ± SD (Web) | ICC(2, 1) | Bias (Web − Native) [t, p-value] | LoA Lower | LoA Upper |
|---|---|---|---|---|---|---|
| CJV (a.u.) | 1.126 ± 0.348 | 1.127 ± 0.346 | 0.9997 | 0.0015 (t = +0.83, p = 0.417) | −0.0149 | 0.0179 |
| CNR (a.u.) | 0.847 ± 0.253 | 0.846 ± 0.252 | 0.9999 | −0.0016 (t = −1.85, p = 0.080) | −0.0092 | 0.0061 |
| EFC (a.u.) | 0.626 ± 0.108 | 0.626 ± 0.108 | 1.0000 | 0.0000 (t = −, p = 1.000) | 0.0000 | 0.0000 |
| FBER (a.u.) | 10338.63 ± 12688.49 | 10338.57 ± 12688.47 | 1.0000 | −0.0582 (t = −0.97, p = 0.345) | −0.5983 | 0.4819 |
| FWHM-avg (mm) | 3.431 ± 0.668 | 3.431 ± 0.668 | 1.0000 | 0.0000 (t = +0.18, p = 0.859) | 0.0000 | 0.0000 |
| FWHM-x (mm) | 3.012 ± 0.610 | 3.012 ± 0.610 | 1.0000 | 0.0000 (t = +1.37, p = 0.187) | 0.0000 | 0.0000 |
| FWHM-y (mm) | 3.921 ± 0.846 | 3.921 ± 0.846 | 1.0000 | 0.0000 (t = +1.59, p = 0.127) | 0.0000 | 0.0000 |
| FWHM-z (mm) | 3.360 ± 0.573 | 3.360 ± 0.573 | 1.0000 | 0.0000 (t = −1.44, p = 0.166) | 0.0000 | 0.0000 |
| SNR-CSF (a.u.) | 1.728 ± 0.201 | 1.728 ± 0.202 | 0.9999 | 0.0003 (t = +0.38, p = 0.709) | −0.0065 | 0.0071 |
| SNR-GM (a.u.) | 5.181 ± 1.461 | 5.186 ± 1.463 | 0.9999 | 0.0053 (t = +1.02, p = 0.319) | −0.0412 | 0.0518 |
| SNR-Total (a.u.) | 6.276 ± 1.332 | 6.278 ± 1.336 | 0.9999 | 0.0025 (t = +0.73, p = 0.475) | −0.0281 | 0.0330 |
| SNR-WM (a.u.) | 11.918 ± 2.896 | 11.920 ± 2.904 | 1.0000 | 0.0018 (t = +0.33, p = 0.748) | −0.0489 | 0.0525 |
| WM2MAX (a.u.) | 0.562 ± 0.211 | 0.562 ± 0.211 | 1.0000 | 0.0000 (t = −0.46, p = 0.650) | −0.0005 | 0.0004 |

**Table 1.** *Bland–Altman agreement between native MRIQC and WebMRIQC on the matched BraTS-Africa cohort (n = 20). Means and standard deviations are reported for each platform. ICC(2,1) is the two-way random-effects, absolute-agreement intraclass correlation coefficient (single measures); values ≥ 0.90 indicate excellent agreement. Bias is the mean difference (Web − Native); t and the p-value are from a paired t-test of the differences against zero ($H_0$: bias = 0). LoA, 95% limits of agreement (bias ± 1.96 SD). ICC(2,1) = 1.000, p = 1.000 and LoA = [0.0000, 0.0000] denote metrics computed identically by both platforms (zero variance in the differences).*

# 5 Discussion & Limitation

Our preliminary findings support the central premise that the MRIQC engine can be delivered through a browser without compromising the integrity of its metrics. Equivalence was strongest for the contrast- and noise-based descriptors (CJV, CNR, SNR and EFC), which are IQMs most informative for routine structural QC. This distinction is important, because it locates the variability in the metrics themselves rather than in the web delivery.

The broader and primary contribution of this platform is open accessibility, by removing local installation, BIDS formatting, and the need for institutional high-performance computing infrastructure and know-how. Its major advantage is in democratizing access to high quality and comparable brain scans, through benchmarking of locally adapted scan parameters from established acquisition protocols, publications, and open datasets. This enables local users in resource-constrained centers to not only obtain the same open-access standardized QC report that has for decades been globally available, but now rapidly understand metrics and their impacts, without specialist training. While the normative percentile comparison and multicentre view features work well for single-subject analyses, efforts are underway to expand these benchmarking functionalities for group-level analyses including for test-rest and longitudinal studies. Beyond providing a basis for MRI harmonization planning, the benchmarking feature is equally intended to support acquisition protocol validation in these regions and generation of regionally adapted benchmarked datasets, where shared acquisition standards can be defined against a common, transparent baseline. More importantly, WebMRIQC can be used as a phantom-based quality control tracker of scanner performance/drift, particularly across Africa where much older scanners and new original equipment manufacturer (OEM) systems are routinely used in clinical service and neuroimaging research. Together, these extended functionalities, advance sustainable imaging capacity in Africa and other LMICs [13].

An important current limitation is that HIPAA compliance and equivalent regional data-protection frameworks have not been fully evaluated or implemented. WebMRIQC is therefore intended strictly as a research tool for de-identified data obtained under institutional approval, and is not validated or approved for clinical diagnostic use or the processing of protected health information; formal regulatory compliance (e.g., HIPAA, GDPR, NDPR) remains part of the future work. Another limitation in this implementation is the inconsistent DICOM metadata that could be encountered by some users, largely due to lack of standard imaging protocols, particularly those tailored to LMIC and emerging clinical scanner configurations. Two of the six images used to test the functionality of the platform from the 6 users (African radiographers), failed DICOM to BIDS conversion and required manual intervention to complete the conversion. The failures were a result of non-standard SeriesDescription and ProtocolName fields in the DICOM tag, which dcm2bids relies on to identify and label series that did not match the expected conventions. The inconsistent naming conventions of the DICOM metadata were not compliant with the requirements of the dcm2bids module, causing the images to result in errors and consequently non-complaint with MRIQC as it requires any input data to be BIDS-compliant. We are exploring avenues to address this in the long-term including creating an internal DICOM meta-data library from de-identified sample datasets, starting with scanners in Africa. In the interim, issues when encountered will be resolved for users through support mechanisms (email contact or AI chatbot).

Although the design and preliminary evaluation of the platform are promising, its dependence on internet connectivity to the shared server and the availability of the server, can pose access challenges. Currently, task prioritization operates on a capacity-bounded First-In First-Out (FIFO) model rather than a weighted multi-tenant scheduler. Considerations for offline and/or edge deployments [24] are being explored for future platform upgrades to address limited internet connectivity. Another consideration is the use of transient storage to protect privacy and reduce storage capacity, but at the expense of prompt result retrieval. The possibility for output data loss due to storage hardware failure, operating or browser system crashes and other software glitches are unavoidable and common challenges of web-based tools. In this evidence, users can retry their analyses as a workaround, while better solutions are explored. We are expanding this initial validation across a larger cohort, including the full BraTS-Africa dataset (n=146) [14] and a normative multi-center African brain MRI dataset (n=350). This will confirm the metric equivalence of our web-based to native implementation. We also note that usability and performance evaluation of the platform was not performed in this preliminary study. Standardized usability measures (e.g., the System Usability Scale), task-completion and time-on-task rates, per-subject processing times, processing failure rates, and controlled workload (concurrent-use) testing will be collected as these structured evaluations for the proposed large-scale validation of WebMRIQC. Finally, the platform is currently oriented to single-subject review, with future updates for group-level harmonization in the works.

## 6 Conclusion and Impact in Resource-constrained settings

This work introduced WebMRIQC, a browser-based, containerized platform that makes standardized MRI quality control accessible in resource-constrained settings. The preliminary evaluation presented here suggests that lowering the barrier to rigorous, harmonized QC, supports equitable participation in collaborative neuroimaging research and provides a practical foundation for regionally adapted quality benchmarks. We intend to use the platform as the validation tool for standardizing neuroimaging acquisition protocols in African populations, specifically for adapting and refining established protocols. To support its wider adoption, we will leverage the strong network of LMIC MRI radiographers, through our Scan with Me (SWiM) training program, to deploy, further validate, and improve the application.

**Acknowledgments.** This study was supported by The Neuro Open Science and Global Health Platform (NeurON), partly funded by the Digital Research Alliance of Canada Research Platforms and Portals (RPP) program. The authors gratefully acknowledge participants of the Scan with Me (SWiM) 2025 Program whose contributions to real-world validation of the platform provided invaluable feedback on its research applicability. The authors also thank Dr Oscar Esteban (University of Lausanne), the developer of MRIQC, for providing the foundational quality-control framework on which WebMRIQC is built, and for his generosity in sharing expertise and feedback that shaped the direction of this work.

**Disclosure of Interests.** The authors have no competing interests to declare that are relevant to the content of this article.

## References

1. Mortamet, B., Bernstein, M. A., Jack, C. R. J., Gunter, J. L., Ward, C., Britson, P. J., Meuli, R., Thiran, J.-P., & Krueger, G. (2009). Automatic quality assessment in structural brain magnetic resonance imaging. *Magnetic Resonance in Medicine*, *62*(2), 365–372. https://doi.org/10.1002/mrm.21992

2. Esteban, O., Birman, D., Schaer, M., Koyejo, O. O., Poldrack, R. A., & Gorgolewski, K. J. (2017). MRIQC: Advancing the automatic prediction of image quality in MRI from unseen sites. *PloS One*, *12*(9), e0184661. https://doi.org/10.1371/journal.pone.0184661

3. Esteban, O., Blair, R. W., Nielson, D. M., Varada, J. C., Marrett, S., Thomas, A. G., Poldrack, R. A., & Gorgolewski, K. J. (2019). Crowdsourced MRI quality metrics and expert quality annotations for training of humans and machines. Scientific data, 6(1), 30. https://doi.org/10.1038/s41597-019-0035-4

4. Hagen, M. P., Provins, C., MacNicol, E., Li, J. K., Gomez, T., Garcia, M., Seeley, S. H., Legarreta, J. H., Norgaard, M., Bissett, P. G., Poldrack, R. A., Rokem, A., & Esteban, O. (2026). Quality assessment and control of unprocessed anatomical, functional and diffusion MRI of the human brain using MRIQC. Nature protocols, 10.1038/s41596-026-01352-y. Advance online publication. https://doi.org/10.1038/s41596-026-01352-y

5. Gorgolewski, K. J., Auer, T., Calhoun, V. D., Craddock, R. C., Das, S., Duff, E. P., Flandin, G., Ghosh, S. S., Glatard, T., Halchenko, Y. O., Handwerker, D. A., Hanke, M., Keator, D., Li, X., Michael, Z., Maumet, C., Nichols, B. N., Nichols, T. E., Pellman, J., Poline, J. B., … Poldrack, R. A. (2016). The brain imaging data structure, a format for organizing and describing outputs of neuroimaging experiments. Scientific data, 3, 160044. https://doi.org/10.1038/sdata.2016.44

6. Montalba, C., Aduluwa, H., Gracia, M. F., Botwe, F., Maharjan, S., Mumuni, A. N., Zeraii, A., Thairu, J., Zurakat‑Aderibigbe, S., Adewo, A., Mokoena, K., Alfaro, A., Flores, G., Lim, T., Mach, M., Adebimpe, A., Taylor, R., Anosike, C., Anyanwu, B., Poloni, G., … Anazodo, U. (2025). Enhancing Dementia Imaging in Low‑and‑Middle Income Countries Through Training of Skilled MRI Personnel. Alzheimer's & Dementia, 21(Suppl 8), e109998. https://doi.org/10.1002/alz70862_109998

7. Adewole, M., Rudie, J. D., Gbadamosi, A., Zhang, D., Raymond, C., Toyobo, O., Omidiji, O., Akinola, R., Suwaid, M. A., Daji, F., Emegoakor, A., Aguh, K., Ojo, N., Kalaiwo, C., Babatunde, G., Ogunleye, A., Gbadamosi, Y., Iorpagher, K., Onuwaje, M., Betiku, B., … BraTS Organizers (2026). Brain tumor segmentation in Sub-Saharan Africa patient population: The BraTS-Africa challenge. Neuro-oncology advances, 8(1), vdag082. https://doi.org/10.1093/noajnl/vdag082

8. Baid, U., Ghodasara, S., Bilello, M., Mohan, S., Calabrese, E., Colak, E., Farahani, K., Kalpathy-Cramer, J., Kitamura, F.C., Pati, S., Prevedello, L.M., Rudie, J.D., Sako, C., Shinohara, R.T., Bergquist, T., Chai, R., Eddy, J.A., Elliott, J., Reade, W.C., Schaffter, T., Yu, T., Zheng, J., Annotators, B., Davatzikos, C., Mongan, J.T., Hess, C.P., Cha, S., Villanueva-Meyer, J.E., Freymann, J.B., Kirby, J.S., Wiestler, B., Crivellaro, P.S., R.Colen, R., Kotrotsou, A., Marcus, D., Milchenko, M., Nazeri, A., Fathallah-Shaykh, H.M., Wiest, R., Jakab, A., Weber, M., Mahajan, A., Menze, B.H., Flanders, A.E., & Bakas, S. (2021). The RSNA-ASNR-MICCAI BraTS 2021 Benchmark on Brain Tumor Segmentation and Radiogenomic Classification. ArXiv, abs/2107.02314.

9. Gorgolewski, K. J., Alfaro-Almagro, F., Auer, T., Bellec, P., Capotă, M., Chakravarty, M. M., Churchill, N. W., Cohen, A. L., Craddock, R. C., Devenyi, G. A., Eklund, A., Esteban, O., Flandin, G., Ghosh, S. S., Guntupalli, J. S., Jenkinson, M., Keshavan, A., Kiar, G., Liem, F., Raamana, P. R., … Poldrack, R. A. (2017). BIDS apps: Improving ease of use, accessibility, and reproducibility of neuroimaging data analysis methods. PLoS computational biology, 13(3), e1005209. https://doi.org/10.1371/journal.pcbi.1005209

10. Li, X., Morgan, P. S., Ashburner, J., Smith, J., & Rorden, C. (2016). The first step for neuroimaging data analysis: DICOM to NIfTI conversion. Journal of neuroscience methods, 264, 47–56. https://doi.org/10.1016/j.jneumeth.2016.03.001

11. Keshavan, A., Yeatman, J. D., & Rokem, A. (2019). Combining Citizen Science and Deep Learning to Amplify Expertise in Neuroimaging. Frontiers in neuroinformatics, 13, 29. https://doi.org/10.3389/fninf.2019.00029

12. Fernandez-Lozano, S., Dadar, M., Morrison, C., Manera, A., Andrews, D., Rajabli, R., Madge, V., St-Onge, E., Shafiee, N., Livadas, A., Fonov, V., Collins, D. L., & Alzheimer's Disease Neuroimaging Initiative (2024). QRATER: a collaborative and centralized imaging quality control web-based application. Aperture neuro, 4, 10.52294/001c.118616. https://doi.org/10.52294/001c.118616

13. Anazodo, U. C., Ng, J. J., Ehiogu, B., Obungoloch, J., Fatade, A., Mutsaerts, H. J. M. M., Secca, M. F., Diop, M., Opadele, A., Alexander, D. C., Dada, M. O., Ogbole, G., Nunes, R., Figueiredo, P., Figini, M., Aribisala, B., Awojoyogbe, B. O., Aduluwa, H., Sprenger, C., Wagner, R., … Consortium for Advancement of MRI Education and Research in Africa (CAMERA) (2023). A framework for advancing sustainable magnetic resonance imaging access in Africa. NMR in biomedicine, 36(3), e4846. https://doi.org/10.1002/nbm.4846

14. Adewole, M., Rudie, J. D., Gbadamosi, A., Zhang, D., Raymond, C., Ajigbotoshso, J., Toyobo, O., Aguh, K., Omidiji, O., Akinola, R., Suwaid, M. A., Emegoakor, A., Ojo, N., Kalaiwo, C., Babatunde, G., Ogunleye, A., Gbadamosi, Y., Iorpagher, K., Onuwaje, M., … Anazodo, U. C. (2025). The BraTS-Africa Dataset: Expanding the Brain Tumor Segmentation Data to Capture African Populations. Radiology: Artificial Intelligence, 7(4), e240528. https://doi.org/10.1148/ryai.240528

15. Udeh-Momoh, C. T., Maina, R., Anazodo, U. C., Akinyemi, R., Atwoli, L., Baker, L., Bassil, D., Blackmon, K., Bosire, E., Chemutai, G., Crivelli, L., Eze, L. U., Ibanez, A., Kafetsouli, D., Karikari, T. K., Khakali, L., Kumar, M., Lengyel, I., de Jager Loots, C. A., Mangialasche, F., … Africa‐FINGERS Study Team (2024). Dementia risk reduction in the African context: Multi-national implementation of multimodal strategies to promote healthy brain aging in Africa (the Africa-FINGERS project). Alzheimer's & dementia : the journal of the Alzheimer's Association, 20(12), 8987–9003. https://doi.org/10.1002/alz.14344

16. Docker Inc. (2026). Docker Desktop (Version 27.0) [Computer software]. https://www.docker.com

17. Boré, A., Guay, S., Bedetti, C., Meisler, S., & GuenTher, N. (2023). Dcm2Bids (Version 3.1.1) [Computer software]. https://doi.org/10.5281/zenodo.8436509

18. Medical Artificial Intelligence Lab (MAI LAB) (2026). https://mailab.io

19. Atkinson, D., Hill, D. L., Stoyle, P. N., Summers, P. E., & Keevil, S. F. (1997). Automatic correction of motion artifacts in magnetic resonance images using an entropy focus criterion. IEEE transactions on medical imaging, 16(6), 903–910. https://doi.org/10.1109/42.650886

20. Tustison, N. J., Avants, B. B., Cook, P. A., Zheng, Y., Egan, A., Yushkevich, P. A., & Gee, J. C. (2010). N4ITK: improved N3 bias correction. IEEE transactions on medical imaging, 29(6), 1310–1320. https://doi.org/10.1109/TMI.2010.2046908

21. Stöcker, T., Keenan, K. E., Knoll, F., Priovoulos, N., Uecker, M., & Zaitsev, M. (2025). Reproducibility and quality assurance in MRI. *Magnetic Resonance Materials in Physics, Biology and Medicine*, *38*(3), 347–352. https://doi.org/10.1007/s10334-025-01271-1

22. Nadal, S., Jovanovic, P., Bilalli, B., & Romero, O. (2022). Operationalizing and automating Data Governance. Journal of big data, 9(1), 117. https://doi.org/10.1186/s40537-022-00673-5

23. Mandl, K. D., & Kohane, I. S. (2015). Federalist principles for healthcare data networks. Nature biotechnology, 33(4), 360–363. https://doi.org/10.1038/nbt.3180

24. Cordova-Cardenas, R., Amor, D., & Gutiérrez, Á. (2025). Edge AI in Practice: A Survey and Deployment Framework for Neural Networks on Embedded Systems. *Electronics*, *14*(24), 4877. https://doi.org/10.3390/electronics14244877